\documentstyle[fullname,palatino,psfig,program]{nemlap}
\normalbaroutside
\begin{document}

\title{Treatment of $\epsilon$-Moves in Subset Construction}

\author{Gertjan van Noord}
\institute{Alfa-informatica \& BCN\\
        University of Groningen, Netherlands\\
        {\tt vannoord@let.rug.nl}
        }
\maketitle

\begin{abstract}
  The paper discusses the problem of determinising finite-state
  automata containing large numbers of $\epsilon$-moves.  Experiments
  with finite-state approximations of natural language grammars often
  give rise to very large automata with a very large number of
  $\epsilon$-moves.  The paper identifies three subset construction
  algorithms which treat $\epsilon$-moves. A number of experiments has
  been performed which indicate that the algorithms differ considerably in
  practice. Furthermore, the experiments suggest that the average number of
  $\epsilon$-moves per state can be used to predict which algorithm is
  likely to perform best for a given input automaton.
\end{abstract}

\section{Introduction}

In experimenting with finite-state approximation techniques for
context-free and more powerful grammatical formalisms (such as the
techniques presented in \namecite{pereira-wright}, \namecite{NE97},
\namecite{grimley}) we have found that the resulting automata often
are extremely large.  Moreover, the automata contain many
$\epsilon$-moves ({\em jumps\/}). And finally, if such automata are
determinised then the resulting automata are often {\em smaller\/}. It
turns out that a straightforward implementation of the subset
construction determinisation algorithm performs badly for such inputs.

As a motivating example, consider the definite-clause grammar that has
been developed for the OVIS2 Spoken Dialogue System. This grammar is
described in detail in \namecite{nle}.  After removing the feature
constraints of this grammar, and after the removal of the sub-grammar
for temporal expressions, this context-free skeleton grammar was input
to an implementation of the technique described in \namecite{NE97}.
\footnote{A later implementation by Nederhof (p.c.) avoids
  construction of the complete non-determistic automaton by
  determinising and minimising subautomata before they are embedded
  into larger subautomata.}
The resulting non-deterministic automaton (labelled {\em zovis2\/}
below) contains 89832 states, 80935 $\epsilon$-moves, and 80400
transitions. The determinised automaton contains only 6541 states, and
60781 transitions.  Finally, the minimal automaton contains only 78
states and 526 transitions!  Other grammars give rise to similar
numbers. Thus, the approximation techniques yield particularly
`verbose' automata for relatively simple languages.

The experiments were performed using the FSA Utilities toolkit
\cite{fsa-two}. At the time, an old version of the toolkit was used,
which ran into memory problems for some of these automata.  For this
reason, the subset construction algorithm has been re-implemented,
paying special attention to the treatment of $\epsilon$-moves. Three
variants of the subset construction algorithm are identified which
differ in the way $\epsilon$-moves are treated:

\begin{description}
\item[per graph] The most obvious and straightforward approach is
  sequential in the following sense. Firstly, an equivalent automaton
  without $\epsilon$-moves is constructed for the input. In order to
  do this, the transitive closure of the graph consisting of all
  $\epsilon$-moves is computed. Secondly, the resulting automaton is
  then treated by a subset construction algorithm for $\epsilon$-free
  automata.
\item[per state] For each {\em state\/} which occurs in a subset
  produced during subset construction, compute the states which are
  reachable using $\epsilon$-moves. The results of this computation
  can be memorised, or computed for each state in a preprocessing
  step. This is the approach mentioned briefly in
  \namecite{instruction-computation}.\footnote{According to Derick
    Wood (p.c.), this approach has been implemented in several
    systems, including Howard Johnson's INR system. }
\item[per subset] For each {\em subset\/} $Q$ of states which arises during
  subset construction, compute $Q' \supseteq Q$ which
  extends $Q$ with all states which are reachable from any member of
  $Q$ using $\epsilon$-moves. Such an algorithm is described in
  \namecite{aho-sethi-ullman}. We extend this algorithm by memorising
  the $\epsilon$-closure computation. 
\end{description}

The motivation for this paper is the experience that the first
approach turns out to be impractical for automata with very large
numbers of $\epsilon$-moves.  An integration of the subset
construction algorithm with the computation of $\epsilon$-reachable
states performs much better in practice.  The {\em per subset\/}
algorithm almost always performs better than the {\em per state\/}
approach. However, for automata with a low number of jumps, the {\em
  per graph\/} algorithm outperforms the others.

In constructing an $\epsilon$-free automaton the number of transitions
increases. Given the fact that the input automaton already is
extremely large (compared to the simplicity of the language it
defines), this is an undesirable situation. An equivalent
$\epsilon$-free automaton for the example given above results in an
automaton with 2353781 transitions. The implementation of {\em per
  subset\/} is the only variant which succeeds in determinising the
input automaton of this example.

In the following section some background information concerning the
{\em FSA Utilities\/} tool-box is provided. Section~\ref{ssc} then
presents a short statement of the problem (determinise a given
finite-state automaton), and a subset construction algorithm which
solves this problem in the absence of $\epsilon$-moves.
Section~\ref{three} identifies three variants of the subset
construction algorithm which take $\epsilon$-moves into account.
Finally, section~\ref{exp} discusses some experiments in order to
compare the three variants both on randomly generated automata and
on automata generated by approximation algorithms.

\section{FSA Utilities}
The {\em FSA Utilities\/} tool-box is a collection of tools to manipulate
regular expressions, finite-state automata and finite-state
transducers (both string-to-string and string-to-weight transducers).
Manipulations include determinisation (both for finite-state acceptors
and finite-state transducers), minimisation, composition,
complementation, intersection, Kleene closure, etc.  
Various visualisation tools are available to browse finite-state
automata. The tool-box is implemented in SICStus Prolog.

The motivation for the {\em FSA Utilities\/} tool-box has been the
rapidly growing interest for finite-state techniques in {\em
  computational linguistics\/}. The {\em FSA Utilities\/} tool-box has
been developed to experiment with these techniques.  The tool-box is
available free of charge under Gnu General Public
License.\footnote{See {\tt http://www.let.rug.nl/\%7Evannoord/Fsa/}.
  The automata used in the experiments are available from the same
  site.}  The following provides an overview of the functionality of
the tool-box.

\begin{itemize}
\item Construction of finite automata on the basis of regular
  expressions. Regular expression operators include concatenation,
  Kleene closure, union and option (the standard regular expression
  operators). Furthermore the extended regular expression operators
  are provided: complement, difference and intersection. Symbols can
  be intervals of symbols, or the `Any'-variable which matches any
  symbol. Regular expression operators are provided for operations on
  the underlying automaton, including minimisation and
  determinisation. Finally, we support user-defined regular expression
  operators.

\item We also provide operators for transductions such as composition,
  cross-product, same-length-cross-product, domain, range, identity
  and inversion. 

\item Determinisation and Minimisation.  Three different minimisation
  algorithms are supported: Hopcroft's algorithm \cite{hopcroft},
  Hopcroft and Ullman's algorithm \cite{hopcroft-ullman}, and
  Brzozowski's algorithm \cite{brzozowski}.

\item Determinisation and minimisation of string-to-string and string-to-weight
  transducers \cite{mohri-det,mohri-cl}. 
  
\item Visualisation.  Support includes built-in visualisation (Tcl/Tk,
  LaTeX, Postscript) and interfaces to third party
  graph visualisation software (Graphviz (dot), VCG, daVinci).

\item Random generation of finite automata (an extension of the algorithm in
  \namecite{leslie} to allow the generation of finite automata
  containing $\epsilon$-moves). 
\end{itemize}

\section{Subset Construction}
\label{ssc}
\subsection{Problem statement}
Let a finite-state machine $M$ be specified by a tuple
$(Q,\Sigma,\delta,S,F)$ where $Q$ is a finite set of states, $\Sigma$
is a finite alphabet, $\delta$ is a function from $Q \times (\Sigma
\cup \{\epsilon\}) \rightarrow 2^Q$.  Furthermore, $S\subseteq Q$ is a
set of start states\footnote{Note that a set of start states is
  required, rather than a single start state. Many operations on
  automata can be defined somewhat more elegantly in this way.
  Obviously, for deterministic automata this set should be a singleton
  set.} and $F\subseteq Q$ is a set of final states.

Let $\epsilon$-move be the relation $\{(q_i,q_j) | q_j \in
\delta(q_i,\epsilon)\}$. $\epsilon$-reachable is the reflexive and
transitive closure of $\epsilon$-move. Let $\epsilon$-CLOSURE: $2^Q
\rightarrow 2^Q$ be a function which is defined as:
\[
\epsilon\mbox{-CLOSURE}(Q') = \{q | q'\in Q', (q',q)\in \epsilon\mbox{-reachable}\}
\]

For any given finite-state automaton $M=(Q,\Sigma,\delta,S,F)$ there
is an equivalent deterministic automaton
$M'=(2^Q,\Sigma,\delta',\{Q_0\},F')$. $F'$ is the set of all states in
$2^Q$ containing a final state of $M$, i.e., the set of subsets
$\{Q_i\in 2^Q| q\in Q_i, q\in F\}$. $M'$ has a single start state
$Q_0$ which is the epsilon closure of the start states of $M$, i.e.,
$Q_0=\epsilon\mbox{-CLOSURE}(S)$.  Finally,

\[
\delta'(\{q_1,q_2,\dots,q_i\},a) = \epsilon\mbox{-CLOSURE}(\delta(q_1,a) \cup \delta(q_2,a)
\cup \dots \cup \delta(q_i,a))
\]

An algorithm which computes $M'$ for a given $M$ will only need to
take into account states in $2^Q$ which are reachable from the start
state $Q_0$. This is the reason that for many input automata the
algorithm does not need to treat all subsets of states (but note that
there are automata for which all subsets are relevant, and hence
exponential behaviour cannot be avoided in general).

\begin{figure}
\begin{center}
\begin{minipage}{.75\textwidth}
\begin{program}
\FUNCT |subset_construction|((Q,\Sigma,\delta,S,F))
  |index_transitions()|; |Trans| := \emptyset; |Finals| := \emptyset; |States| := \emptyset;
  |Start| =: |epsilon_closure|(S)
  |add|(Start)
  \WHILE \mbox{ there is an unmarked subset } T\in |States| \DO
  |mark|(T)
  \FOREACH (a,U) \in |instructions|(T) \DO
     U := |epsilon_closure|(U)
     |Trans|[T,a] := \{U\}
     |add|(U)
    \OD
    \OD
  \keyword{return} ~(|States|,\Sigma,|Trans|,\{|Start|\},|Finals|)
\END
\mbox{   }
\PROC |add|(U)\rcomment{Reachable-state-set Maintenance}
\IF U \notin |States| 
\THEN 
    \mbox{ add }U\mbox{ unmarked to }|States|
    \IF U\cap F \THEN |Finals| := |Finals|\cup U\FI
\FI
\END
\mbox{   }
\FUNCT |instructions|(P)\rcomment{Instruction Computation}
\keyword{return} ~|merge|(\bigcup_{p\in P} |transitions|(p))
\END
\mbox{   }
\FUNCT |epsilon_closure|(U) \rcomment{variant 1: No $\epsilon$-moves}
\keyword{return} ~U
\END
\end{program}
\end{minipage}
\end{center}
\caption{\label{subset-c}Subset-construction algorithm.}
\end{figure}

Consider the subset construction algorithm in figure~\ref{subset-c}.
The algorithm maintains a set of subsets \(|States|\). Each subset can be
either marked or unmarked (to indicate whether the subset has been treated
by the algorithm); the set of unmarked subsets is sometimes referred to
as the agenda. The algorithm takes such an unmarked subset $T$ and
computes all transitions leaving $T$. This computation is performed by 
the function \(|instructions|\) and is called {\em instruction
  computation} by \namecite{instruction-computation}. 

The function \(|index_transitions|\) constructs the function
\(|transitions|: Q \rightarrow \Sigma\times 2^Q \).  This function
returns for a given state $p$ the set of pairs $(s,T)$ representing
the transitions leaving $p$.  Furthermore, the function \(|merge|\)
takes such a set of pairs and merges all pairs with the same first
element (by taking the union of the corresponding second elements).
For example:

\begin{program} 
|merge|(\{(a,\{1,2,4\}), (b,\{2,4\}), (a,\{3,4\}), (b,\{5,6\})\}) = \{(a,\{1,2,3,4\}),(b,\{2,4,5,6\}) \} 
\end{program}

The procedure \(|add|\) is responsible for `reachable-state-set
maintenance', by ensuring that target subsets are added to the
set of subsets if these subsets were not encountered before.
Moreover, if such a new subset contains a final state, then this
subset is added to the set of final states.

\section{Three Variants for $\epsilon$-Moves}
\label{three}
The algorithm presented in the previous section does not treat
$\epsilon$-moves. In this section three possible extensions of the
algorithm are identified to treat $\epsilon$-moves.

\subsection{Per graph}
This variant can be seen as a straightforward implementation of the
constructive proof that for any given automaton with $\epsilon$-moves
there is an equivalent one without $\epsilon$-moves
\cite{hopcroft-ullman}[page 26-27].

For a given $M=(Q,\Sigma,\delta,S,F)$ this variant first computes 
$M'=(Q,\Sigma,\delta',S',F)$, where
$S'=\epsilon\mbox{-CLOSURE}(S)$, and 
$\delta'(q,a) = \epsilon\mbox{-CLOSURE}(\delta(q,a))$.  The function
$\epsilon\mbox{-CLOSURE}$ is computed by using a standard transitive
closure algorithm for directed graphs: this algorithm is applied to
the directed graph consisting of all $\epsilon$-moves of $M$. Such an
algorithm can be found in several textbooks (see, for instance,
\namecite{algorithms}). 

The advantage of this approach is that the subset construction
algorithm does not need to be modified at all.  Moreover, the
transitive closure algorithm is fired only once (for the full graph),
whereas the following two variants call a specialised transitive
closure algorithm possibly many times. 

\subsection{Per subset and per state} 
The {\em per subset\/} and the {\em per state\/} algorithm use a variant
of the transitive closure algorithm for graphs. Instead of computing
the transitive closure of a given graph, this algorithm only computes
the closure for a given set of states. 
Such an algorithm is given in figure\ref{closure}.

\begin{figure}
\begin{center}
\begin{minipage}{.75\textwidth}
\begin{program}
\FUNCT |closure|(T)
D =: \emptyset
\FOREACH t\in T \DO \mbox{ add } t \mbox{ unmarked to } D\OD
\WHILE \mbox{ there is an unmarked state } t\in D \DO
    |mark|(t)
    \FOREACH q\in\delta(t,\epsilon) \DO
    \IF q\notin D \THEN \mbox{ add } q \mbox{ unmarked to } D \FI
    \OD
\OD
\keyword{return} ~D
\END
\end{program}
\end{minipage}
\end{center}
\caption{\label{closure}Epsilon-closure Algorithm}
\end{figure}

In either of the two integrated approaches, the subset construction
algorithm is initialised with an agenda containing a single subset
which is the $\epsilon$-CLOSURE of the set of start-states of the
input; furthermore, the way in which new transitions are computed also 
takes the effect of $\epsilon$-moves into account. Both differences
are accounted for by an alternative definition of the
\(|epsilon_closure|\) function. 

The approach in which the transitive closure is computed for one state
at a time is defined by the following definition of the
\(|epsilon_closure|\) function. Note that we make sure that the
transitive closure computation is only performed once for each input
state, by memorising the \(|closure|\) function.

\begin{program}
\FUNCT |epsilon_closure|(U) \rcomment{variant 2: per state}
\keyword{return}~ \bigcup_{u\in U}|memo|(|closure|(\{u\}))
\END
\end{program}

In the case of the {\em per subset\/} approach the closure algorithm
is applied to each subset. We also memorise the closure function, in
order to ensure that the closure computation is performed only once
for each subset. This can be useful since the same subset can be
generated many times during subset construction. The definition simply is:

\begin{program}
\FUNCT |epsilon_closure|(U) \rcomment{variant 3: per subset}
\keyword{return}~ |memo|(|closure|(U))
\END
\end{program}

The motivation for {\em per state\/} approach may be the insight that
in this case the closure algorithm is called at most $|Q|$ times. In
contrast, in the {\em per subset\/} approach the transitive closure
algorithm may need to be called $2^{|Q|}$ times. On the other hand, in
the {\em per state\/} approach some overhead must be accepted for
computing the union of the results for each state. Moreover, in
practice the number of subsets is often much smaller than $2^{|Q|}$.
In some cases, the number of reachable subsets is smaller than the
number of states encountered in those subsets.

\section{Experiments}
\label{exp}
Two sets of experiments have been performed. In the first set of
experiments a number of random automata is generated according to a
number of criteria (based on \namecite{leslie}) . In the second set of
experiments, results are provided for a number of (much larger)
automata that surfaced during actual development work on finite-state
approximation techniques.

\paragraph{Random automata.} 
Firstly, consider a number of experiments for randomly generated
automata.  Following \namecite{leslie}, the {\em absolute transition
  density\/} of an automaton is defined as the number of transitions
divided by the square of the number of states times the number of
symbols (i.e. the number of transitions divided by the number of
possible transitions).  {\em Deterministic transition density\/} is
the number of transitions divided by the number of states times the
number of symbols (i.e. the ratio of the number of transitions and the
number of possible transitions in a deterministic machine).
\namecite{leslie} shows that {\em deterministic transition density\/} is a
reliable measure for the difficulty of subset construction.
Exponential blow-up can be expected for input automata with
deterministic transition density of around 2.\footnote{Leslie uses the
  terms {\em absolute density\/} and {\em deterministic density\/}.}

\begin{figure}
\centerline{\psfig{file=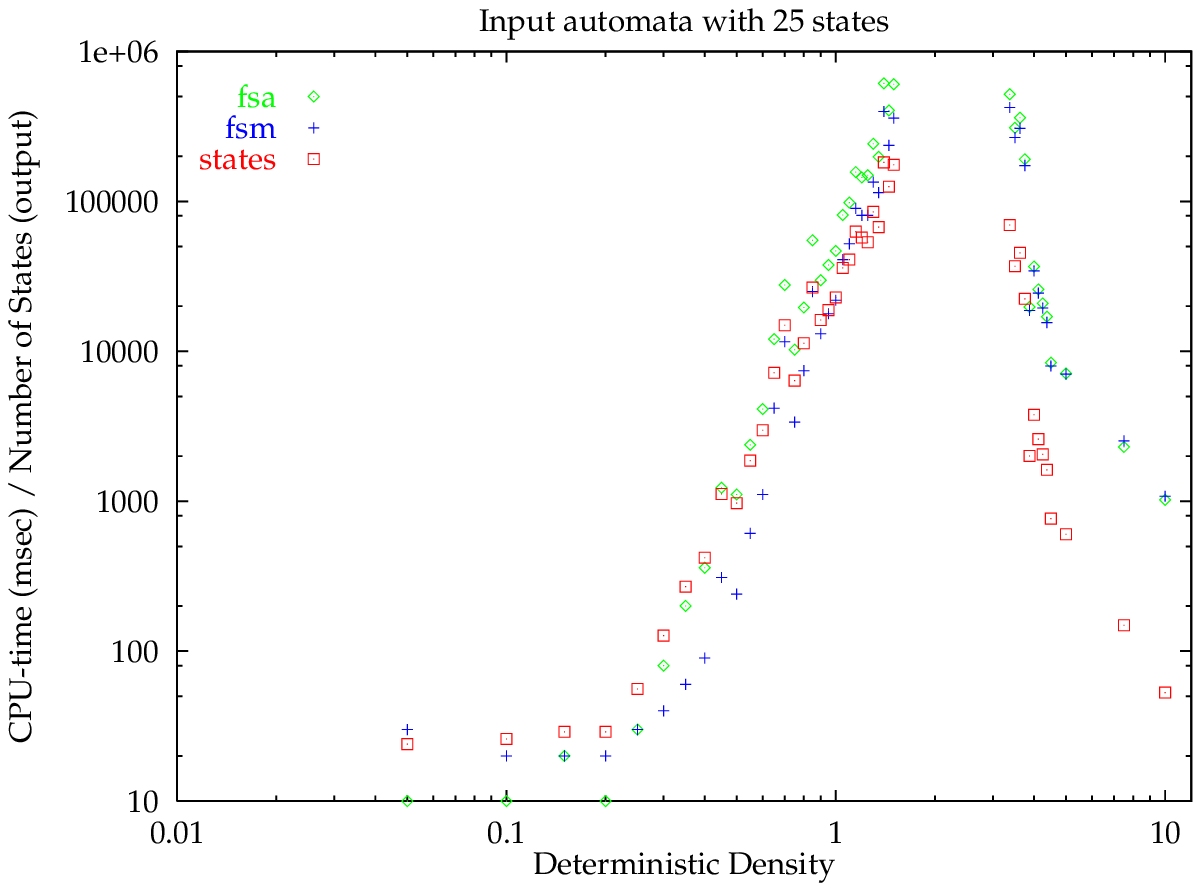,width=\textwidth}}

\hspace{1ex}

\centerline{\psfig{file=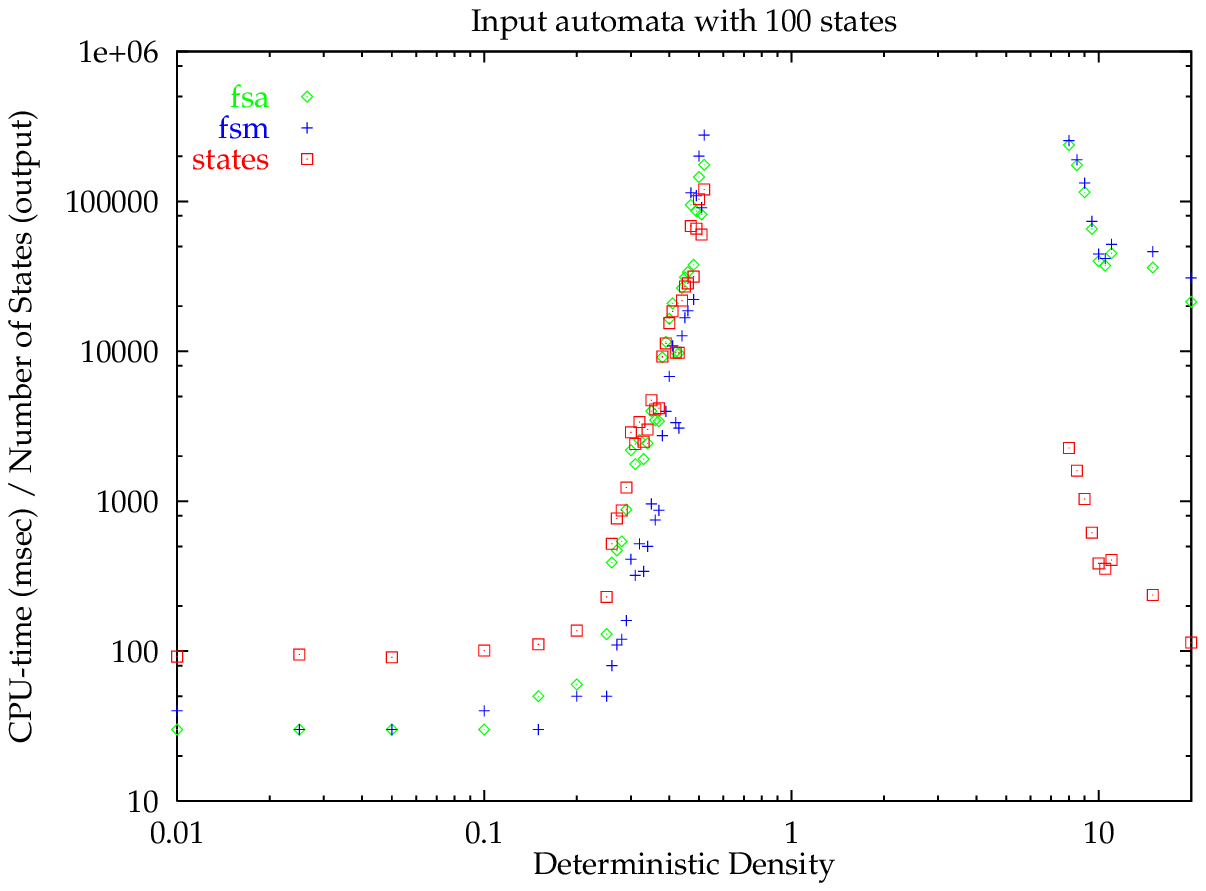,width=\textwidth}}
\caption{\label{expone}Deterministic transition density versus
  CPU-time in msec. The input automata have no $\epsilon$-moves.}
\end{figure}

A number of automata were generated randomly, according to
the number of states, symbols, and transition density.  The random
generator makes sure that all states are reachable from the start
state.  For the first experiment, a number of automata was randomly
generated, consisting of 15 symbols, and 15, 20, 25, 100 or 1000
states, using various densities (and no $\epsilon$-moves). The results
are summarised in figure~\ref{expone}.  Only a single result is given
since each of the implementations works equally well in the absence of
$\epsilon$-moves.\footnote{CPU-time was measured on a HP 9000/780
  machine running HP-UX 10.20, 240Mb, with SICStus Prolog 3 \#3. For
  comparison with an ``industrial strength'' implementation, we have
  applied the determiniser of AT\&T's FSM utilities for the same
  examples. The results show that for automata with very small
  transition densities FSM is faster (up to 2 or 3 times as fast), but
  for automata with larger densities the results are very similar, in
  some cases our Prolog implementation is even faster. Note finally
  that our timings do include IO, but not the start-up of the Prolog
  engine. }

\begin{figure}
\centerline{\psfig{file=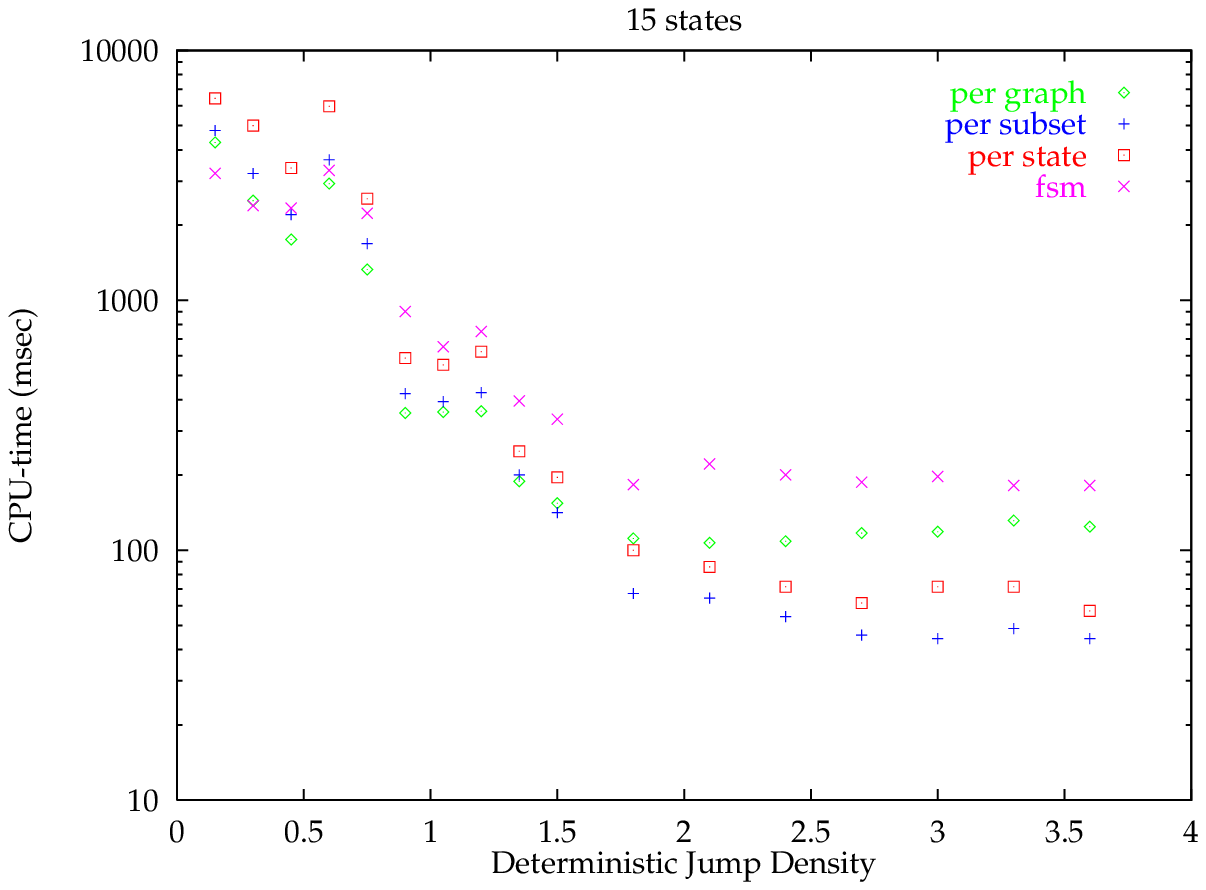}}
\caption{\label{exptwo}Average amount of CPU-time versus jump density
  for each of the three algorithms, and FSM. Input automata have 15
  states. Absolute transition densities: 0.01-0.3. }
\end{figure}

\begin{figure}
\centerline{\psfig{file=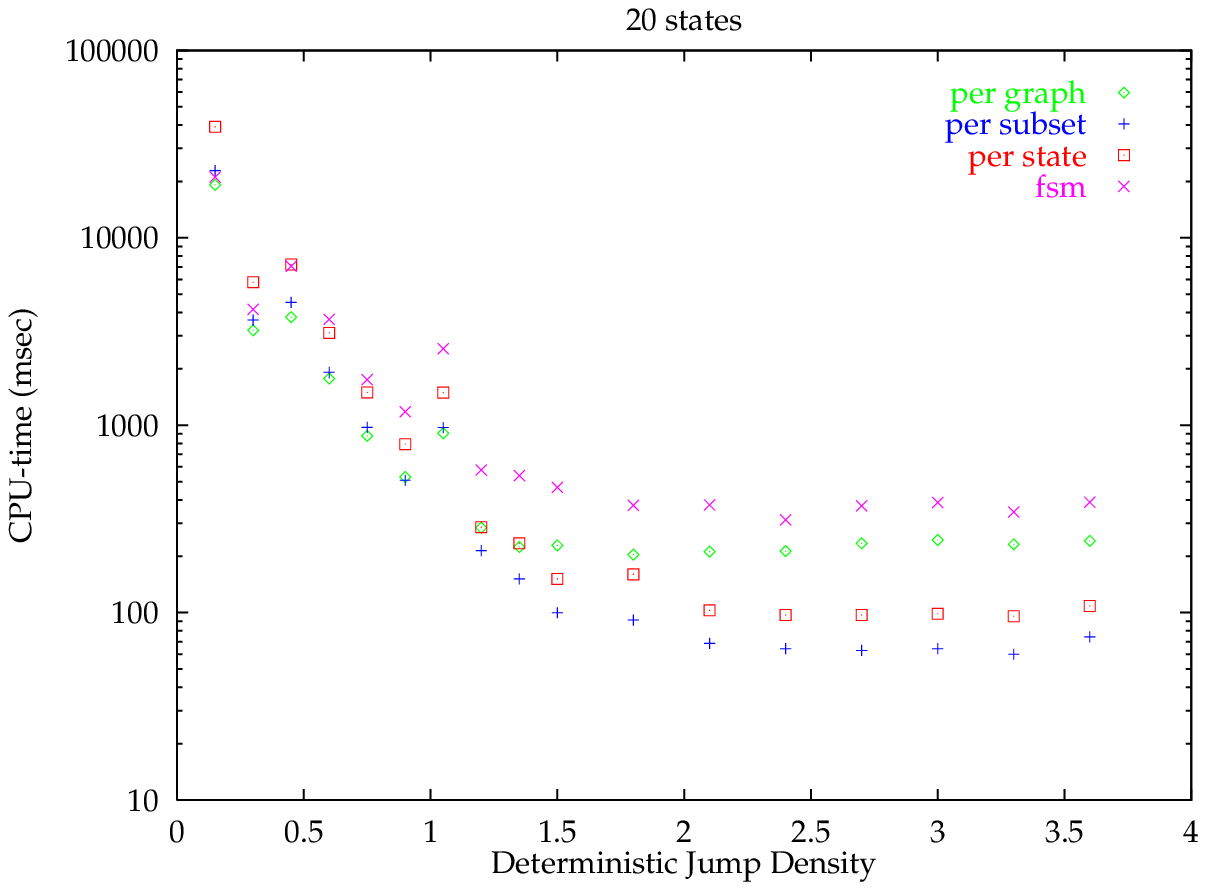}}
\caption{\label{expthree}Average amount of CPU-time versus jump density
  for each of the three algorithms, and FSM. Input automata have 20
  states. Absolute transition densities: 0.01-0.3. }
\end{figure}

\begin{figure}
\centerline{\psfig{file=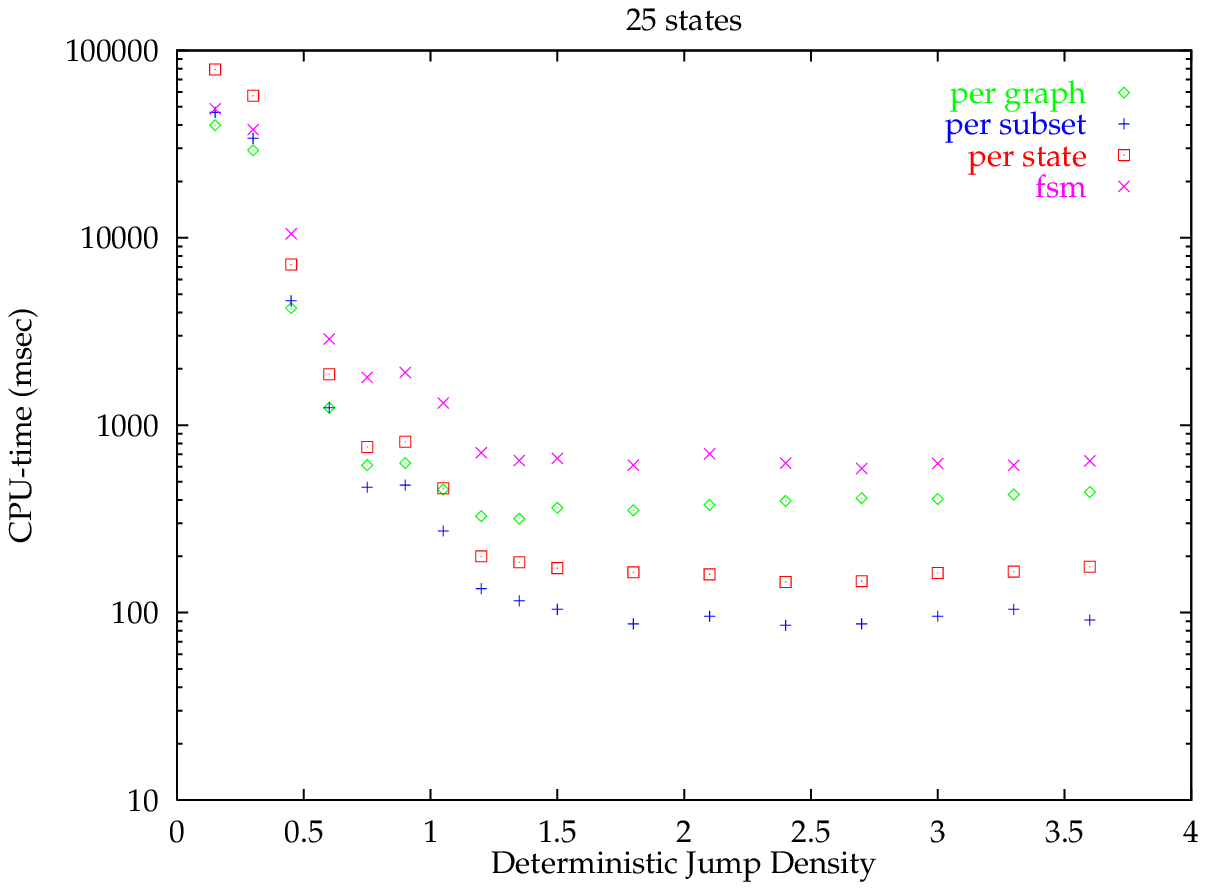}}
\caption{\label{expfour}Average amount of CPU-time versus
  deterministic jump density
  for each of the three algorithms, and FSM. Input automata have 25
  states. Absolute transition densities: 0.01-0.3.}
\end{figure}

\begin{figure}
\centerline{\psfig{file=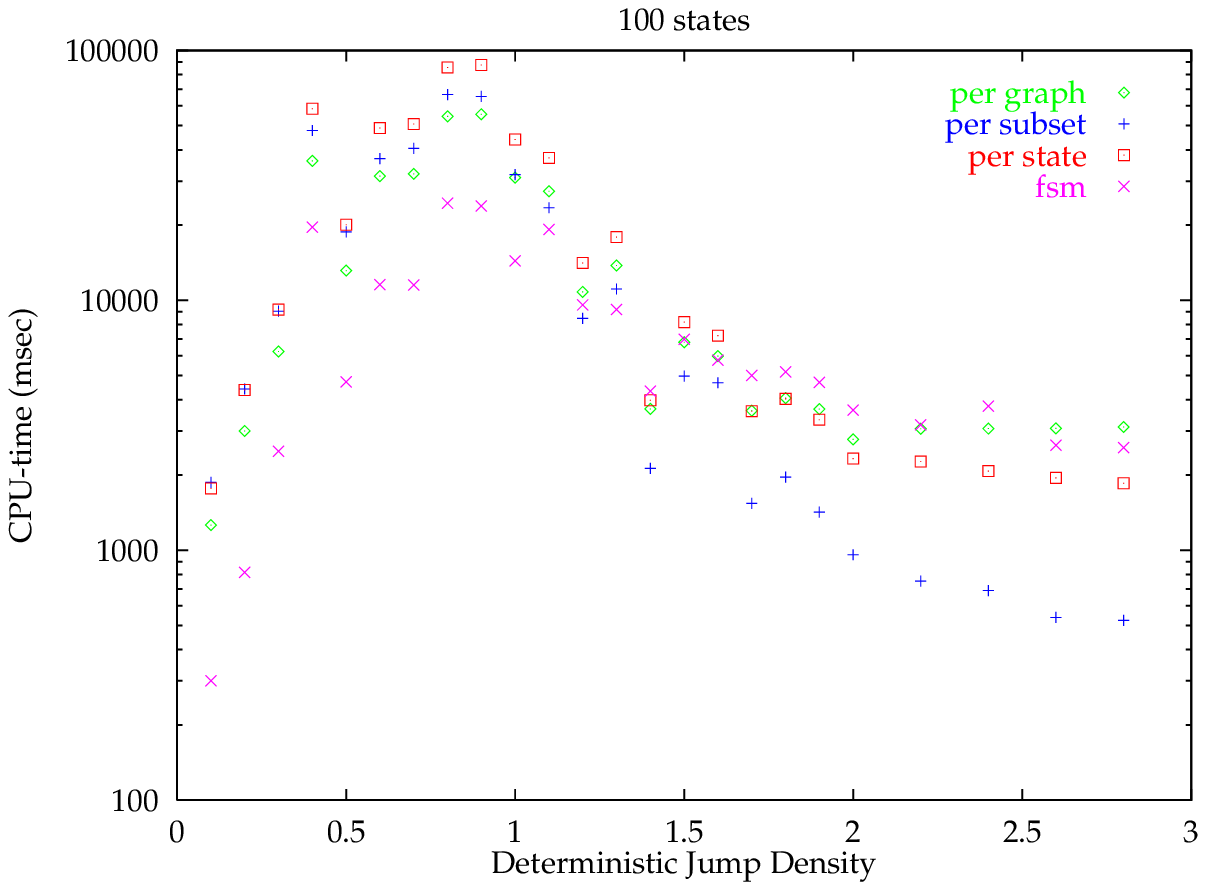}}
\caption{\label{expfive}Average amount of CPU-time versus
  deterministic jump density for each of the three algorithms, and FSM.  Input
  automata have 100 states. Absolute transition densities:
  0.001-0.0035. }
\end{figure}

A new concept called {\em absolute jump density\/} is introduced to
specify the number of $\epsilon$-moves. It is defined as the number of
$\epsilon$-moves divided by the square of the number of states (i.e.,
the probability that an $\epsilon$-move exists for a given pair of
states).  Furthermore, {\em deterministic jump density\/} is the
number of $\epsilon$-moves divided by the number of states (i.e., the
average number of $\epsilon$-moves which leave a given state).  In
order to measure the differences between the three implementations, a
number of automata has been generated consisting of 15 states and 15
symbols, using various transition densities between 0.01 and 0.3 (for
larger densities the automata tend to collapse to an automaton for
$\Sigma^*$). For each of these transition densities, jump densities
were chosen in the range 0.01 to 0.24 (again, for larger values the
automaton collapses). In figure~\ref{exptwo} the outcomes of this
experiment are summarised by listing the average amount of CPU-time
required per deterministic jump density (for each of the three
algorithms). Thus, every dot represents the average for determinising
a number of different input automata with various absolute transition
densities and the same deterministic jump density.  The
figures~\ref{expthree}, \ref{expfour} and \ref{expfive} summarise
similar experiments using input automata with 20, 25 and 100
states.\footnote{We also provide the results for FSM again; we used
  the pipe {\tt fsmrmepsilon | fsmdeterminize }. According to Fernando
  Pereira (pc) the comparison is less meaningful in this case because
  the fsmrmepsilon program treats weighted automata. This generalisation
  requires some overhead also in case no weights are used (for the
  determiniser this generalisation does not lead to any significant
  overhead). Pereira mentions furthermore that FSM used to include a
  determiniser with integrated treatment of jumps. Because this version
  could not (easily) be generalised for weighted automata it was
  dropped from the tool-set.}

The striking aspect of these experiments is that the {\em per graph\/}
algorithm is more efficient for lower deterministic jump densities,
whereas, if the deterministic jump density gets larger, the {\em per
  subset\/} algorithm is more efficient. The turning point is around a
deterministic jump density between 1 and 1.5, where it seems that for
larger automata the turning point occurs at a lower determinisic jump
density. Interestingly, this generalisation is supported by the
experiments on automata which were generated by approximation
techniques (although the results for randomly generated automata are
more consistent than the results for `real' examples).

\paragraph{Experiment: Automata generated by approximation algorithms}
The automata used in the previous experiments were randomly generated,
according to a number of criteria. However, it may well be that in
practice the automata that are to be treated by the algorithm have
typical properties which were not reflected in this test data. For
this reason results are presented for a number of automata that were
generated using approximation techniques for context-free grammars
\cite{pereira-wright,NE97,grimley}. In particular, a number of
automata has been used generated by Mark-Jan Nederhof using the
technique described in \namecite{NE97}. In addition, a small number of
automata have been used which were generated using the technique of
\namecite{pereira-wright} (as implemented by Nederhof).

The automata typically contain lots of jumps. Moreover, the number of
states of the resulting automaton is often {\em smaller\/} than the
number of states in the input automaton. Results are given in
table~\ref{tableone}. One of the most striking examples is the {\em
  ygrim} automaton consisting of 3382 states and 10571 jumps. For this
example, the {\em per graph\/} implementation ran out of memory (after a
long time), whereas the {\em per subset\/} algorithm produced the
determinised automaton relatively quickly. The FSM implementation took 
much longer for this example (whereas for many of the other examples
it performs better than our implementations). Note that this example
has the highest number of jumps per number of states ratio.

\begin{table}
\centerline{
\begin{tabular}{|r|rrr|rrrr|}\hline
 & \multicolumn{3}{|c}{input automaton} &
\multicolumn{4}{|c|}{CPU-time (msec)} \\
Id & \#states & \# transitions & \# jumps & graph & subset & state & {\em FSM}
 \\\hline
griml.n &238    &43     &485    &2060   &100    &140    &40\\
g9a     &342    &58     &478    &260    &70     &70     &30\\
g7      &362    &424    &277    &180    &240    &200    &60\\
g15     &409    &90     &627    &280    &130    &180    &40\\
ovis5.n &417    &702    &130    &290    &320    &380    &190\\
g9      &438    &313    &472    &560    &850    &640    &110\\
g11     &822    &78     &1578   &1280   &160    &160    &60\\
g8      &956    &2415   &330    &500    &500    &610    &140\\
g14     &1048   &403    &1404   &1080   &1240   &730    &120\\
ovis4.n &1424   &2210   &660    &2260   &2220   &2870   &1310\\
g13     &1441   &1006   &1404   &2400   &3780   &2550   &440\\
rene2   &1800   &2597   &96     &440    &530    &600    &200\\
ovis9.p &1868   &2791   &3120   &83340  &80400  &87040  &52560\\
ygrim   &3382   &5422   &10571  &-      &2710   &70140  &784910\\
ygrim.p &48062  &63704  &122095 &-      &1438960&-      &8575850\\
java19  &54369  &28333  &59394  &130130 &55290  &64420  &8470\\
java16  &64210  &43935  &43505  &67180  &24200  &31770  &6370\\
zovis3  &88156  &78895  &79437  &-      &968160 &-      &768440\\
zovis2  &89832  &80400  &80935  &-      &1176650&-      &938040\\
\hline
\end{tabular}
}
\caption{\label{tableone}Results for automata generated by
  approximation algorithms. The dashes in the table indicate
  that the corresponding algorithm ran out of memory (after a long
  period of time) for that
  particular example. }
\end{table}

\section{Conclusion}
We have discussed three variants of the subset-construction algorithm
for determinising finite automata. The experiments support the
following conclusions:

\begin{itemize}
\item the {\em per graph\/} variant works best for automata with a
  limited number of jumps
\item the {\em per subset\/} variant works best for automata with a
  large number of jumps
\item the {\em per state\/} variant almost never outperforms both of the two
  other variants
\item typically, if the deterministic jump density of the input is
  less than 1, then the {\em per graph\/} variant outperforms the {\em per
    subset\/} variant. If this value is larger than 1.5, then {\em per
    subset\/} outperforms {\em per graph\/}.
\item the {\em per subset\/} approach is especially useful for automata
  generated by finite-state approximation techniques, because those
  techniques often yield automata with very large number of
  $\epsilon$-moves.
\end{itemize}

\section*{Acknowledgements}
I am grateful to Mark-Jan Nederhof for support, and for providing me
with lots of (often dreadful) automata generated by his finite-state
approximation tools.

\small
\bibliography{biblio}

\end{document}